\documentclass[10pt, reqno, letterpaper, twoside]{amsart}
\usepackage[margin=1in]{geometry}
\usepackage{booktabs}
\usepackage{array}
\usepackage{amssymb, bm, mathtools}
\usepackage[usenames,dvipsnames,svgnames,table]{xcolor}
\usepackage[pdftex, xetex]{graphicx}
\usepackage{enumerate, setspace}
\usepackage{float, colortbl, tabularx, longtable, multirow, subcaption, environ, wrapfig, textcomp, booktabs}
\usepackage{pgf, tikz, framed, url, hyperref}
\usepackage[normalem]{ulem}
\usetikzlibrary{arrows,positioning,automata,shadows,fit,shapes}
\usepackage[english]{babel}
\usepackage{verbatim}
\usepackage{microtype}
\microtypecontext{spacing=nonfrench}
\usepackage{times}
\title{LemmaHead: RAG Assisted Proof Generation Using Large Language Models}
\author{
Tianbo Yang [yangti@seas.upenn,edu],
Mingqi Yan [yan77alex@gmail.com],
Hongyi Zhao [hongyizhao12@gmail.com],
Tianshuo Yang [yangts@umich.edu]
}

\begin{document}

\begin{abstract}
Developing the logic necessary to solve mathematical problems or write mathematical proofs is one of the more difficult objectives for large language models (LLMS). Currently, the most popular methods in literature consists of fine-tuning the model on written mathematical content such as academic publications and textbooks, so that the model can learn to emulate the style of mathematical writing. In this project, we explore the effectiveness of using retrieval augmented generation (RAG) to address gaps in the mathematical reasoning of LLMs. We develop LemmaHead, a RAG knowledge base that supplements queries to the model with relevant mathematical context, with particular focus on context from published textbooks. To measure our model's performance in mathematical reasoning, our testing paradigm focuses on the task of automated theorem proving via generating proofs to a given mathematical claim in the Lean formal language.
\end{abstract}

\maketitle

\section{Introduction}

Formal proof generation is one of the foundations of advanced mathematics. Despite the recent advancements in large language models (LLMs), automating the process of generating and verifying formal mathematical proofs remains a significant challenge. Formal proofs are important for verifying the correctness of mathematical reasoning, and automating this process can assist mathematicians, educators, and students in exploring complex problems with greater efficiency and accuracy. Our goal is to enable LLMs to generate formal proofs from mathematical problems while ensuring correctness through validation. By leveraging retrieval-augmented generation (RAG) to provide the model with authoritative mathematical references, we aim to improve reasoning accuracy and facilitate formal proof generation.

\section{Background}

Mathematics, particularly at advanced levels such as those encountered in Olympiad competitions, presents a formidable challenge due to its reliance on rigorous reasoning, precise notation, and creative problem-solving strategies. Traditional computational tools are good at numerical calculations and symbolic manipulations but struggle to handle the detailed logical structures and formal proofs required in high-level mathematics. This limitation underscores the need for systems that can not only understand but also generates and verify formal mathematical proofs.

Recent advances in natural language processing (NLP) and large language models (LLMs) have demonstrated remarkable capabilities in reasoning, contextual understanding, and text generation. Models like GPT-4 have shown promise in solving mathematical problems, but their utility for formal proof generation remains constrained by a lack of structured access to authoritative mathematical knowledge and the challenges of formal proof representation in languages like Lean. Furthermore, the absence of a robust verification mechanism complicates efforts to ensure the validity of generated proofs, which are essential in mathematical reasoning.

To address these challenges, our project leverages retrieval-augmented generation (RAG) to enhance the LLM's ability to generate accurate, contextually grounded mathematical proofs. By integrating authoritative mathematical resources, such as textbooks on algebra, geometry, and number theory, into a structured knowledge base, we aim to provide the model with precise, contextually relevant references. This structured approach not only improves the reasoning quality of the LLM but also bridges the gap between informal mathematical language and formal proof systems like Lean.

\section{Related Work}

Extensive previous research inspired the methodology of this research. From  Ahn et al.(2024)\cite{1}'s review over LLM application over math problem solving, LLM Prompt and In-text example are highlightened for their significant impact over performance. Empirical evidence from other reseach also proposed the significance of relevant in-text context. Didolkar et al.(2024)\cite{2} shows proposes a theoritical framework that LLM performance can be boosted through case specific skills and provide empirical evidence through math problem solving. Inspired by this research, we continued the experiment on providing more relevant context in prompt to improve LLM performance in this scenario. Building on these insights, Dong et al.(2024)\cite{3} integrate reinforcement learning mechanisms with LLMs to enable automated theorem proving, showing the importance of reward-based approaches for proof generation. Frieder et al.(2023)\cite{4} further investigate LLMs like GPT-4 and Claude 2, demonstrating the ability to "search" for relevant information when tackling mathematical problems, bridging gaps between formal systems and LLM reasoning. Lewis et al.(2020)\cite{5} introduce RAG models that combine parametric memory from pre-trained seq2seq models with non-parametric memory accessed through dense vector retrieval, demonstrating its ability to address tasks requiring explicit knowledge access and manipulation. Liu et al.(2023)\cite{6} construct a robust benchmark for evaluating proof generation models, providing a foundation for measuring LLM performance in formal reasoning tasks. Zheng et al.(2021)\cite{7}'s paper provides the miniF2F dataset, which illustrates a unified cross-system benchmark for formal Olympiad-level mathematical problems.

\section{Methodology}

We use OpenAi API to access GPT-4 as our baseline LLM model. In the control group, we directly prompt GPT-4 to generate formal proofs in the Lean language, before running the generated code in Lean to verify its correctness. Our experimental group is divided into three separate pipelines integrating GPT-4 with our LemmaHead RAG knowledge base, allowing it to provide mathematical context for the LLM to use to augment its response. The first pipeline makes a simple query to LemmaHead for mathematical context, while the other two pipelines employ enhanced query generation (EQG) and iterative proof augmentation (IPA) to more fully leverage the benefits provided by LemmaHead. 

For evaluation, we compare the performance of the baseline GPT-4 with our RAG-assisted models.
Our evaluation metric is the rate at which the models successfully generate a correct proof in the Lean formal language. This makes it easy to check the correctness of the generated proofs using a proof verification algorithm.

\subsection{Datasets}

For evaluation, we compare the performance of the baseline GPT-4 with our RAG-assisted models on the MiniF2F dataset. MiniF2F  consists of 488 informal problem statements drawn from the AIME, AMC, and the International Mathematical Olympiad (IMO), as well as material from high-school and undergraduate mathematics courses \cite{7}. These problems are divided equally into a validation set and a test set.

\subsection{Constructing the LemmaHead Retrieval-Augmented Generation (RAG) Knowledge Base}

A key component of our approach involves building a retrieval-augmented generation (RAG) pipeline that supplies the Large Language Model (LLM) with relevant mathematical references. These references are drawn from a corpus of authoritative textbooks commonly used in preparation for Olympiad-level mathematics competitions. The selected textbooks includes essential domain knowledge —inequalities, number theory, algebra, and functional equations—and provide problem statements, theoretical results, example solutions, and methodological insights. Our goal is to incorporate this textual data into a structured RAG database, enabling the model to leverage these foundational materials to improve reasoning quality and answer accuracy.

\subsubsection{Data Extraction and Normalization}

A significant challenge arose during the digitization of these textbooks, as conventional PDF-to-text extraction tools struggled to accurately capture mathematical expressions and special symbols intrinsic to Olympiad-style problem statements and proofs. To address this, we employed a multi-stage pipeline centered around a vision-based approach. First, we segmented the PDFs into individual page images, ensuring each page’s layout and content—especially diagrams, equations, and notations—were preserved. Next, we utilized a GPT-based image recognition capability to read each page image and transcribe the content into a LaTeX-formatted text. By converting visual data directly into LaTeX, we maintained the fidelity of complex equations and structural relationships among mathematical objects, allowing for subsequent processing and embedding with minimal loss of information.

\subsubsection{Chunking and Structuring the Content}

To ensure the RAG database is semantically coherent and contextually meaningful, we carefully designed a chunking strategy. Rather than separating problems from their corresponding solutions—a fragmentation that could dilute the contextual integrity—we preserved these pairs within the same chunk. This way, each chunk reflects a self-contained logical unit, such as a single problem and its solution, or a cohesive theoretical exposition, rather than interleaving multiple unrelated items. To facilitate this, we crafted system prompts guiding the GPT-based parser to split the LaTeX-rendered text into logically distinct segments. Instructions emphasized the identification and preservation of thematic boundaries, ensuring that theoretical discussions, worked examples, and problem-solution pairs were isolated into discrete, context-rich chunks. This segmentation preserves the integral meaning of each source item, making it easier for the retrieval module to surface relevant pieces of information at inference time.

\subsubsection{Populating the RAG Database}

With the textbook content segmented into coherent LaTeX-based chunks, each unit was then embedded using the CHROMA RAG framework. Through GPT-based embedding functions, we transformed each chunk into a vector representation that captures its semantic essence. These embeddings, stored in the CHROMA database, serve as the backbone of our retrieval process. At inference time, when the LLM encounters a new Olympiad-level problem—such as one drawn from an AMC or International Mathematical Olympiad (IMO) scenario—it queries the RAG database. The database returns the top semantically relevant segments, such as proven lemmas, known inequalities, classic examples, or similar problem structures and their respective solution patterns. By incorporating these retrieved references into the prompt, the model gains immediate, contextually grounded knowledge. This provides the LLM with a strong foundation upon which it can reason more effectively, apply established theorems, recall problem-solving strategies, and ultimately improve its performance on challenging mathematical tasks.

\subsection{Constructing the RAG pipelines}

\begin{figure}[htb]
  \centering
  \includegraphics[width=0.4\textwidth]{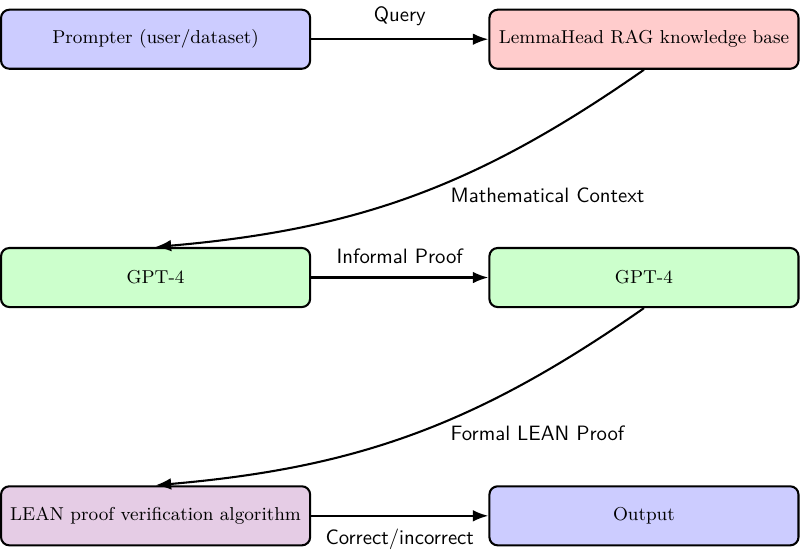}
  \caption{Basic RAG pipeline}
  \label{fig:experiment1}
\end{figure}

With the RAG knowledge base constructed, we can now integrate it with GPT-4 and Lean to complete our automated proof generation pipeline. The basic process is shown in figure \ref{fig:experiment1}.

For each problem in the MiniF2F dataset, we obtain its informal problem statement (in LaTex) and use it to generate a query to LemmaHead. LemmaHead then returns the relevant mathematical context based on a similarity search using word embeddings. Next, we combine informal problem statement with the retrieved context to get an augmented prompt to GPT-4. GPT-4 is first prompted to generate an informal proof (in latex), and then asked to convert the informal proof into a formal Lean proof. Finally, we run the the generated formal proof in the Lean environment to verify its correctness.

\subsubsection{Enhanced query generation (EQG)}

To better utilize the LemmaHead knowledge base, we apply enhanced query generation (EQG). In the RAG pipeline with EQG, we first prompt GPT-4 to generate a list of keywords of mathematical concepts, theorems, lemmas, and propositions needed to solve the problem as described by the informal problem statement. Using these keywords, we produce an enhanced query to LemmaHead that targets mathematical knowledge relevant to the problem. This enhanced query drastically improves the quality and comprehensiveness of the context returned from LemmaHead.

\begin{figure}[htb]
  \centering
  \includegraphics[width=0.4\textwidth]{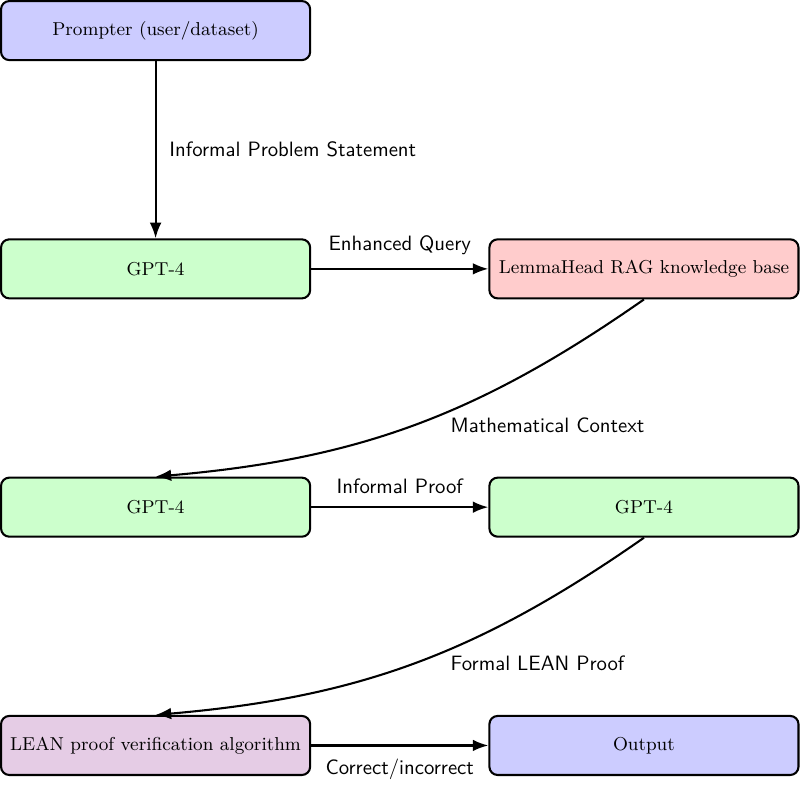}
  \caption{RAG pipeline with enhanced query generation}
  \label{fig:experiment2}
\end{figure}

Our second RAG pipeline tests the impact of EQG on the overall task of RAG assisted proof generation. The process is illustrated in figure \ref{fig:experiment2}.

\subsubsection{Iterative proof augmentation (IPA)}

Both the basic RAG framework and RAG with EQG rely on GPT-4's ability to perform zero-shot context retrieval and proof generation. However, our experiments demonstrate that the LLM's ability to perform either task is highly unreliable. To address this weakness, we propose iterative proof augmentation (IPA), a technique for iteratively improving query and proof generation. Each time after generating an informal proof, we perform EQG by prompting GPT-4 again to generate a list of keywords of mathematical concepts, theorems, lemmas, and propositions needed to solve the problem, with both the original informal problems statement and the newly generated informal proof as context. These keywords are then used to query LemmaHead to provide higher quality context, which is further used by the LLM to write an improved version of the informal proof. The process is repeated for $\sigma$ iterations. 

\begin{figure}[htb]
  \centering
  \includegraphics[width=0.4\textwidth]{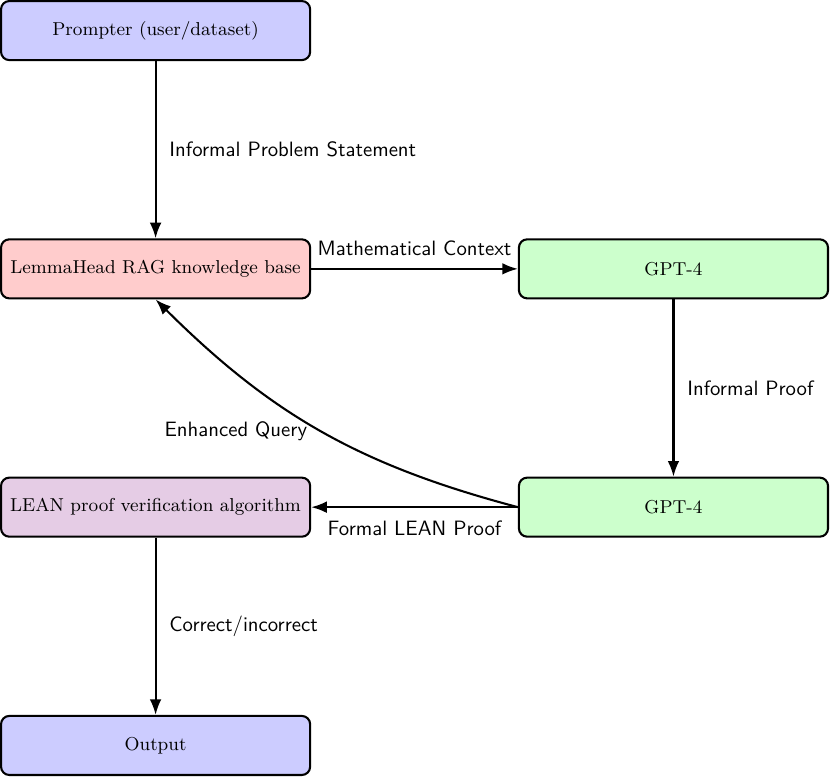}
  \caption{RAG pipeline with iterative proof augmentation}
  \label{fig:experiment3}
\end{figure}

In our experiments, we arbitrarily set $\sigma=5$ and observed significant improvements in the quality and correctness of both informal and formal generated proofs. IPA is the basis for our third RAG pipeline, illustrated in figure \ref{fig:experiment3}.

\section{Experimental Results}

We test our models on both the validation and test datasets of MiniF2F. The problems in MiniF2F are given as informal statements in LaTex form \cite{7}, so no pre-processing is needed. We measure the Pass@1 rate of each model, which is the correctness rate of the generated formal Lean proof after only 1 attempt by the model. Thus, no model can try multiple times to generate a proof. 

Table \ref{tab:validation} displays the models' performances in the MiniF2F validation set, while table \ref{tab:test} shows their correctness rates on the test set. Besides our models, we included the performances of two state-of-the-art LLM proof generators for comparison. Human-guided GPT-4 is a model whose context is supplemented with human written informal proofs \cite{6}, while GPT-$f$ is a GPT-3 model finetuned on large quantities of mathematical data \cite{7}.

\begin{table}[htb]
\centering
\begin{tabular}{|c|c|c|c|c|c|}
\hline
GPT-4 & RAG-assisted GPT-4 & RAG-assisted GPT-4 with EQG & RAG-assisted GPT-4 with IPA & Human-guided GPT-4 & GPT-$f$  \\ \hline
9.4\%   & 2.3\%   & 25.2\%   & 40.0\%   & 11.5\%   &  23.9\%  \\ \hline
\end{tabular}
\caption{Correctness rates of formal proof generators on the MiniF2F validation dataset \cite{6}, \cite{7}}
\label{tab:validation}
\end{table}

\begin{table}[htb]
\centering
\begin{tabular}{|c|c|c|c|c|c|}
\hline
GPT-4 & RAG-assisted GPT-4 & RAG-assisted GPT-4 with EQG & RAG-assisted GPT-4 with IPA & Human-guided GPT-4 & GPT-$f$  \\ \hline
9.0\%   & 2.5\%   & 27.6\%   & 32.4\%   & 8.6\%   &  24.6\%  \\ \hline
\end{tabular}
\caption{Correctness rates of formal proof generators on the MiniF2F test dataset \cite{6}, \cite{7}}
\label{tab:test}
\end{table}

As shown in the table, RAG-assisted GPT-4 with IPA substantially outperforms the state-of-the-art models in terms of Pass@1 rate on both datasets, while RAG-assisted GPT-4 with EQG slightly outperforms GPT-$f$. An interesting observation is that basic RAG-assisted GPT-4 without any improvements drastically underperforms even compared to GPT-4 without RAG. This may be due to poor querying of LemmaHead resulting in unrelated context being extracted, demonstrating that RAG can actually be detrimental to the performance of LLM proof generators if context retrieval is not handled properly.

\section{Conclusion}

In this paper, we developed LemmaHead, a RAG knowledge base that supplements queries to the model with relevant mathematical context, and demonstrated the potential of RAG-assisted LLM proof automation. With the addition of EQG and IPA, we have shown that RAG-assisted LLM proof generation can substantially outperform state-of-the-art proof generators.

Although our methodology showed promising results relative to state-of-the-art models, we have not addressed the limitations fundamental to current LLM-based automated proof systems. Currently, even the most advanced language models such GPT-4 struggle with the task of zero-shot and few-shot mathematical proof writing, especially for difficult IMO-level problems. There are two deficiencies that comprise this issue; the first is that LLMs often lose track of the crucial information when generating long proofs, resulting in the final output being incomplete. Newer models are much better at generating complete proofs, but they still struggle with the second problem: logically putting together a correct proof. To address these limitations, we proposed IPA, but this technique multiplies the inference time of the model by the number of iterations.

In our experiments, we only tested RAG-assisted proof generation with IPA for a small number (5) of iterations due to computational constraints. Even so, we observed a substantial improvement in Pass@1 correctness rates, and that these rates seem to increase with more iterations. For further research, we would like to test the performance of IPA for large numbers of iterations. We hypothesize that there is a limit to the potential of IPA due to diminishing returns in improvement for increasing the number of iterations. Furthermore, we would also like to explore RAG-assisted proof generation with models other than GPT-4, such as Llama, Gemini, and Claude.

\bibliographystyle{plain} 
\bibliography{reference}

\end{document}